\documentclass[letterpaper, 10 pt, conference]{ieeeconf}  

\usepackage{amsmath}
\usepackage{graphicx}
\usepackage{multirow}
\usepackage{multicol}
\usepackage{booktabs}
\usepackage{amssymb}
\usepackage[table,xcdraw]{xcolor}
\usepackage{microtype}
\usepackage{url}

\def\VspaceL{\vspace{-0.11cm}}
\def\VspaceS{\vspace{-0.08cm}}

\IEEEoverridecommandlockouts                              
\title{\LARGE \bf
D\textsuperscript{3}epth: Self-Supervised Depth Estimation with Dynamic Mask \\ in Dynamic Scenes
}

\author{Siyu Chen \quad Hong Liu$^{*}$ \quad Wenhao Li \quad Ying Zhu \quad Guoquan Wang \quad Jianbing Wu
\thanks{$^{*}$Corresponding author}
\thanks{Siyu Chen, Hong Liu, Wenhao Li, Ying Zhu, Guoquan Wang, and Jianbing Wu are with the State Key Laboratory of General Artificial Intelligence, Peking University, Shenzhen Graduate School, China.
E-mail: {\tt\footnotesize\{csyunling,YingZhu,guoquanwang\}@stu.pku.edu.cn}, {\tt\footnotesize\{hongliu,wenhaoli\}@pku.edu.cn}, {\tt\footnotesize kimbingng@gmail.com}.}
}

\begin{document}

\maketitle
\thispagestyle{empty}
\pagestyle{empty}

\begin{abstract}
Depth estimation is a crucial technology in robotics. 
Recently, self-supervised depth estimation methods have demonstrated great potential as they can efficiently leverage large amounts of unlabelled real-world data. 
However, most existing methods are designed under the assumption of static scenes, which hinders their adaptability in dynamic environments. To address this issue, we present D\textsuperscript{3}epth, a novel method for self-supervised depth estimation in dynamic scenes. 
It tackles the challenge of dynamic objects from two key perspectives. 
First, within the self-supervised framework, we design a reprojection constraint to identify regions likely to contain dynamic objects, allowing the construction of a dynamic mask that mitigates their impact at the loss level. 
Second, for multi-frame depth estimation, we introduce a cost volume auto-masking strategy that leverages adjacent frames to identify regions associated with dynamic objects and generate corresponding masks. This provides guidance for subsequent processes.
Furthermore, we propose a spectral entropy uncertainty module that incorporates spectral entropy to guide uncertainty estimation during depth fusion, effectively addressing issues arising from cost volume computation in dynamic environments. 
Extensive experiments on KITTI and Cityscapes datasets demonstrate that the proposed method consistently outperforms existing self-supervised monocular depth estimation baselines. Code is available at \url{https://github.com/Csyunling/D3epth}.
\end{abstract}

\section{INTRODUCTION}
Depth information is crucial for various real-world applications, such as autonomous vehicles \cite{wang2019pseudo}, robot navigation \cite{biswas2012depth}, and human-robot interaction \cite{flacco2012depth}. Self-supervised depth estimation \cite{ garg2016unsupervised, bae2023deep, scv3} has emerged as a promising method due to its ability to predict pixel-level depth maps from images without requiring annotated data. It infers depth from a single target image and supervises learning by constructing a photometric loss through the reprojection of adjacent images back to the target image \cite{zhou2017unsupervised}. 

\begin{figure}[t]
\centering
\includegraphics[width=1.00\linewidth]{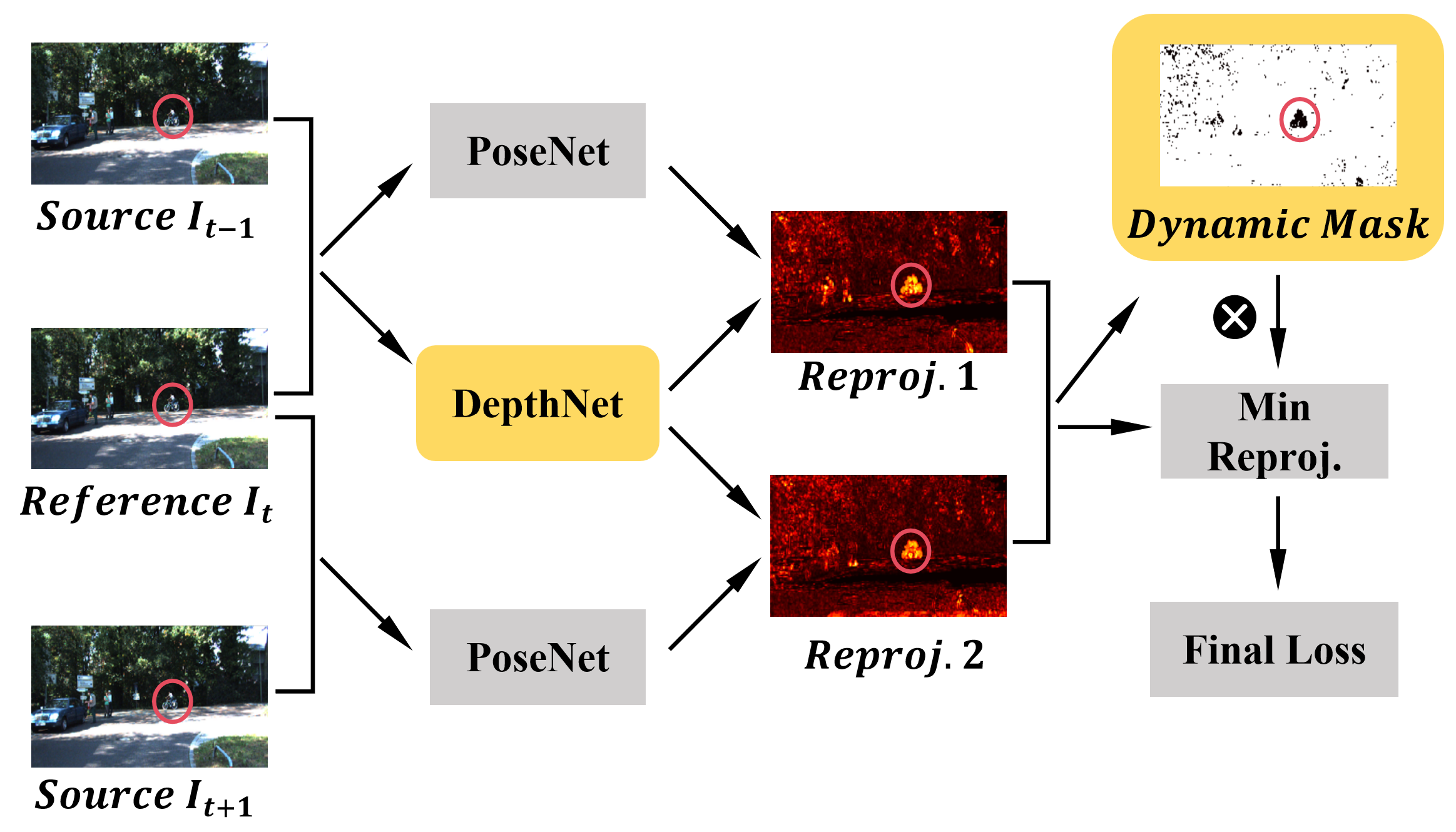}
\caption{
Overall framework of the proposed D\textsuperscript{3}epth. We propose a \textbf{Dynamic Mask (DM)} within the self-supervised framework by masking regions that are likely to be dynamic objects, identified where both reprojection losses exhibit high values. Additionally, we tackle the issue of dynamic objects from the perspective of depth estimation in \textbf{DepthNet}, focusing primarily on multi-frame depth estimation.  
}
\label{overviewall}
\VspaceL
\end{figure}

\begin{figure}[t]
\centering
\includegraphics[width=1.00\linewidth]{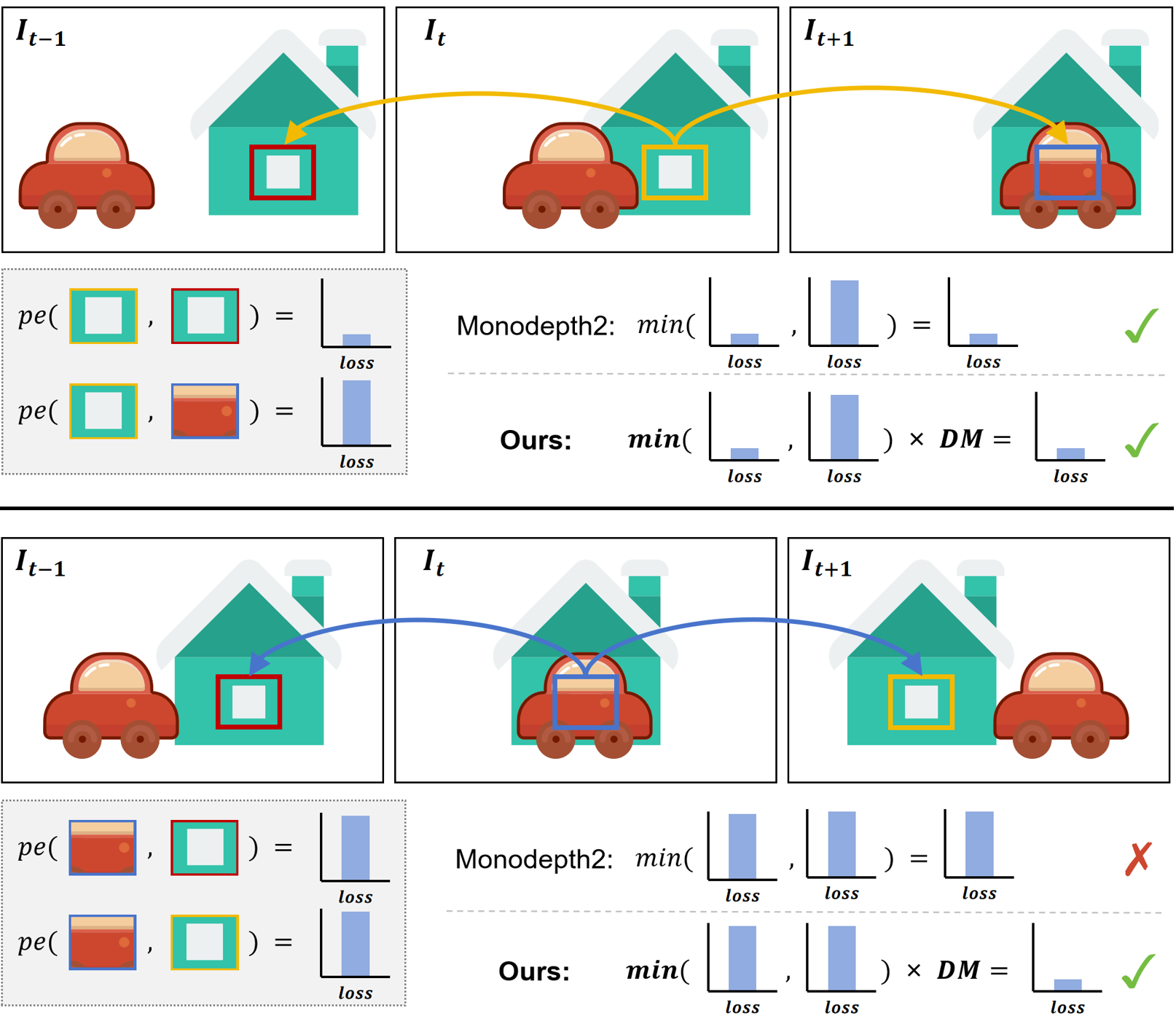}
\caption{\textbf{Comparison of our D\textsuperscript{3}epth with Monodepth2.} The upper part of the figure illustrates the typical occlusion scenario addressed by Monodepth2. Both our method and Monodepth2 can handle this situation. However, the lower part of the figure depicts a case where Monodepth2 fails to resolve, as it relies on the minimum of two reprojection losses. In contrast, our method effectively addresses this issue by performing an additional calculation of a Dynamic Mask (DM) to correct the loss.}
\label{DM}
\VspaceL
\end{figure}

Compared to single-frame methods, multi-frame self-supervised depth estimation \cite{Manydepth, DynamicDepth, DualRefine} has garnered more attention for practical applications such as autonomous navigation, as it leverages richer information by inferring depth maps using multiple images during inference. However, both single-frame and multi-frame self-supervised methods are typically built on the assumption of static scenes, which presents challenges when dynamic objects are encountered in real-world environments. This stems from two primary factors:

\textbf{(i)}
The common framework of self-supervised depth estimation \cite{zhou2017unsupervised} relies on the reprojection loss to capture geometric relationships between consecutive frames under the assumption of photometric consistency. However, this assumption breaks down in the presence of dynamic objects, as shown by the red-circled section in Fig. \ref{overviewall}, where a person riding a bike leads to errors during the view synthesis phase, significantly degrading the accuracy of the generated depth maps. 
To mitigate the impact of dynamic objects, Monodepth2 \cite{monodepth2} proposes the Minimum Reprojection Loss. 
However, this approach is not exhaustive and fails to address the challenging dynamic conditions illustrated in the lower part of Fig. \ref{DM}.

\textbf{(ii)}
In multi-frame depth estimation, the construction of cost volumes does not account for dynamic objects and occlusions, introducing additional errors and making multi-frame depth estimation more susceptible to challenges posed by dynamic scenes. 
To alleviate this, some studies \cite{Manydepth, DualRefine, DynamicDepth} use teacher-student distillation, where a single-frame depth network guides the multi-frame depth estimation network to correct feature matching errors caused by cost volumes. Other methods involve semantic segmentation to identify dynamic objects and adjust their matching \cite{DynamicDepth, MAL} or use optical flow estimation \cite{DS-Depth} to alleviate the impact of dynamic objects. 
Unfortunately, these techniques often introduce complex algorithms with increasing inference costs.

In this paper, we propose \textbf{D\textsuperscript{3}epth} (Self-Supervised \textbf{Depth} Estimation with \textbf{D}ynamic Mask in \textbf{D}ynamic Scenes), a novel method for addressing the challenging problem of dynamic scenes. 
Specifically, we approach the issue from the general self-supervised framework by utilizing the reprojection loss to identify regions likely affected by dynamic objects, thereby establishing a dynamic mask to reduce the impact of dynamic objects at the loss level, without adding additional inference overhead. 
As illustrated in Fig. \ref{overviewall}, for a set of three consecutive frames, we compute two sets of reprojection losses. If both sets exhibit high losses in certain regions (highlighted by the red-circled area where a person is riding a bike), it suggests that these regions are affected by dynamic objects, enabling us to mask them out. Our method builds upon the Minimum Reprojection Loss proposed in Monodepth2 \cite{monodepth2}  and offers a more robust solution for handling occlusions and dynamic objects. Monodepth2 addresses occlusions, as illustrated in the upper part of Fig. \ref{DM}, where one of the adjacent frames is occluded. In this case, taking the minimum of the reprojection losses from two frames helps correct the issue. However, Monodepth2 fails to handle the scenario depicted in the lower part of Fig. \ref{DM}, where a region in the current frame is occluded by a dynamic object, while the corresponding regions in both the previous and subsequent frames remain unobstructed. Our method effectively addresses this case by identifying dynamic regions and correcting the reprojection loss, whereas Monodepth2 struggles with occlusions caused by dynamic objects due to its reliance on the minimum of two high-loss areas.

From the perspective of depth estimation in DepthNet, we focus on multi-frame depth estimation. This approach constructs a cost volume through feature matching to regress depth values, which makes it more susceptible to issues caused by dynamic objects compared to single-frame depth estimation. To address this, we introduce the Cost Volume Auto-Masking strategy, which generates masks for potential dynamic object regions based on adjacent frame images. These masks guide subsequent computations. Furthermore, we integrate spectral entropy after the cost volume to capture richer information for uncertainty estimation and depth fusion, effectively addressing the dynamic challenges introduced by cost volumes.

In summary, the contributions of this paper are as follows:
\begin{itemize}
\item We propose D\textsuperscript{3}epth for self-supervised depth estimation in dynamic scenes. 
It utilizes the Dynamic Mask to address the problem of dynamic objects by masking regions with high reprojection losses, enhancing results without extra computational cost during inference. 
\item A Cost Volume Auto-Masking strategy is introduced to identify dynamic regions before cost volume construction, forming the basis for subsequent processes. Additionally, a Spectral Entropy Uncertainty module is proposed to use these identified regions to guide uncertainty estimation and improve depth fusion. 
\item Our D\textsuperscript{3}epth achieves state-of-the-art results on the Cityscapes and KITTI datasets.
\end{itemize}

\begin{figure*}[t]
\centering
\includegraphics[width=1.00\linewidth]{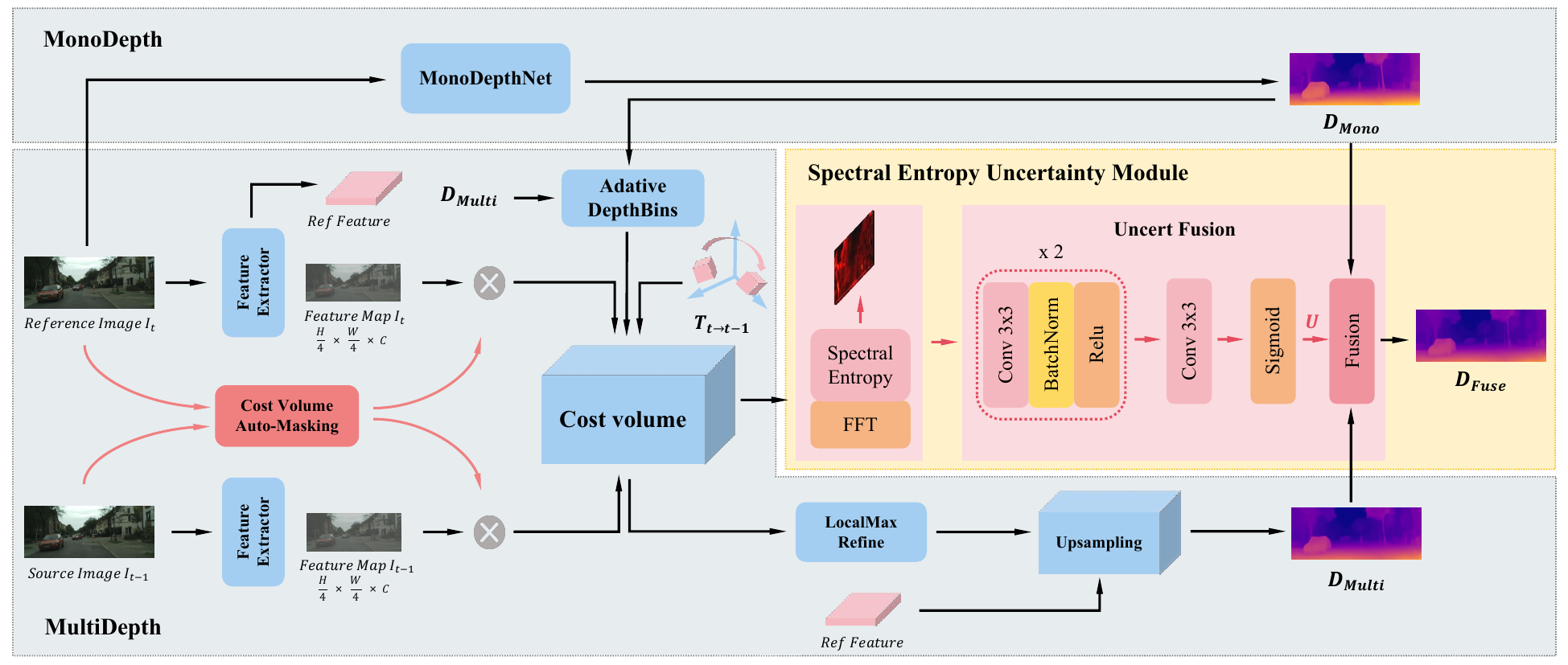}
\caption{Overall framework of \textbf{DepthNet} in our \textbf{D\textsuperscript{3}epth}. It consists of three main modules: single-frame depth estimation (MonoDepth), multi-frame depth estimation (MultiDepth), and Spectral Entropy Uncertainty (SEU) module. 
Cost Volume Auto-Masking is applied before computing the cost volume to filter out regions affected by dynamic objects and to guide subsequent processing. 
And the SEU module leverages the guidance from Cost Volume Auto-Masking and incorporates spectral entropy to enrich the information. This combined approach enhances uncertainty estimation and improves the fusion of MonoDepth and MultiDepth.}
\label{DepthOverview}
\VspaceL
\end{figure*}

\section{RELATED WORK}

\noindent \textbf{Single-Frame Self-Supervised Depth Estimation.}
Self-supervised monocular depth estimation has gained significant attention, owing to its capability to predict pixel-level depth maps from a single image without the need for labeled data. The foundational framework for self-supervised depth estimation is introduced by Zhou et al. \cite{zhou2017unsupervised}, who employ DepthNet and PoseNet to predict geometric relationships between frames. 

Monodepth2 \cite{monodepth2} addresses occlusion issues by proposing the minimum reprojection loss and employs full-resolution multi-scale sampling to mitigate visual artifacts. Further advancements are made by MonoProb \cite{marsal2024monoprob}, which incorporates an interpretable confidence measure to enhance depth modeling through uncertainty estimation. SQLdepth \cite{wang2024sqldepth} introduces a self-supervised method utilizing the Self Query Layer to build a self-cost volume, effectively capturing fine-grained scene geometry from a single image. For indoor depth estimation, PMIndoor \cite{chen2023pmindoor} leverages multiple loss functions to constrain depth estimation in non-textured regions and proposes a pose rectification network to address camera pose inaccuracies. To handle adverse weather conditions, WeatherDepth \cite{wang2024weatherdepth} introduces a curriculum-based contrastive learning strategy, integrating robust weather adaptation techniques to tackle challenging environments without inducing knowledge forgetfulness. 

\noindent \textbf{Multi-Frame Self-Supervised Depth Estimation.}
Compared to single-frame methods, multi-frame depth estimation leverages multiple images for inference, providing richer information. This approach has high practical value in applications such as autonomous driving, making it an increasing focus in recent research. Early approaches \cite{cs2018depthnet, patil2020don} to multi-frame depth estimation often employ Recurrent Neural Networks (RNNs) to enhance model performance; however, these methods typically come with high computational costs. 
ManyDepth \cite{Manydepth} introduces an adaptive cost volume feature matching scheme to exploit geometric information between frames using multiple inputs at test time. Building on this, Guizilini et al. \cite{depthformer} propose a transformer architecture for cost volume generation, improving multi-view feature matching. To address dynamic object challenges, DynamicDepth \cite{DynamicDepth} designs an occlusion-aware cost volume to facilitate geometric reasoning in regions occluded by motion.  MAL \cite{MAL} introduces a motion-aware loss and a distillation scheme to correct errors caused by object motion in multi-frame self-supervised depth estimation. DS-Depth \cite{DS-Depth} employs a dynamic cost volume with residual optical flow to better handle moving objects and occlusions, complemented by a fusion module balancing static and dynamic cost volumes. 

In contrast, our method primarily addresses the dynamic object problem from the loss level. Both the proposed Dynamic Mask and Spectral Entropy Uncertainty modules do not introduce additional inference costs.

\section{METHOD}

\subsection{Preliminary}

Self-supervised methods use reprojection loss between adjacent frames as supervision for training. Each training instance includes a reference frame \( I_t \) and two temporally adjacent source frames \( I_s \) (\( s \in \{t - 1, t + 1\} \)). Given the depth map \( D_t \) of the target view and the camera pose \( T_{t \rightarrow s} \) between the target and source views, the image synthesis from the source to the target view can be formulated as:
\begin{equation}
I_{s \rightarrow t} = I_s\left\langle \text{Proj}(D_t, T_{t \rightarrow s}, K) \right\rangle,
\end{equation}
where \( K \) represents the known camera intrinsics, \( \left\langle \cdot \right\rangle \) is the sampling operator \cite{jaderberg2015spatial}, and \(\text{Proj}(\cdot)\) is the coordinate projection operation \cite{zhou2017unsupervised}. The photometric loss is composed of L1 and SSIM \cite{SSIM}:
\begin{equation}
pe(\boldsymbol{I}_a,\boldsymbol{I}_b)=\frac\alpha2(1-SSIM(\boldsymbol{I}_a,\boldsymbol{I}_b))+(1-\alpha)\left\|\boldsymbol{I}_a-\boldsymbol{I}_b\right\|_1.
\label{photometric}
\end{equation}

Following Monodepth2 \cite{monodepth2}, we adopt the per-pixel minimum reprojection loss as our photometric loss:
\begin{equation}
\mathcal{L}_{ph}(\boldsymbol{I}_t,\boldsymbol{I}_{s\to t})=\min_spe(\boldsymbol{I}_t,\boldsymbol{I}_{s\to t}).
\label{eq:photometric}
\end{equation}

\subsection{Overview} 
Our method is primarily designed to address issues caused by dynamic objects from two perspectives, as shown in Fig. \ref{overviewall}. First, we introduce the Dynamic Mask (DM) (Sec. \ref{subsec:dynamic_mask}) to tackle limitations in the self-supervised framework by analyzing the reprojection loss. 
Second, we propose improvements to DepthNet, focusing on multi-frame depth estimation, as illustrated in Fig. \ref{DepthOverview}. The overall framework of DepthNet consists of three components: single-frame depth estimation (MonoDepth), multi-frame depth estimation (MultiDepth), and the proposed Spectral Entropy Uncertainty (SEU).
In our approach, Cost Volume Auto-Masking is applied before cost volume computation to filter out regions affected by dynamic objects and guide subsequent processing (Sec. \ref{subsec:CVAM}). The SEU module leverages spectral entropy to provide richer information, enhancing uncertainty estimation and enabling more effective depth fusion (Sec. \ref{subsec:SEU}). Further details will be introduced in the following sections.

\subsection{Dynamic Mask}
\label{subsec:dynamic_mask}
In self-supervised depth estimation frameworks, dynamic objects, lighting changes, and occlusions pose significant challenges during the computation of reprojection loss, as these factors violate the photometric consistency assumption. This often leads to errors in reprojection loss calculations, particularly in the affected areas. 

To address this issue, we propose the Dynamic Mask method, which targets this problem at the loss level, as illustrated in Fig. \ref{overviewall}. By analyzing the reprojection errors from two source images, it can be observed that dynamic objects typically correspond to regions with high loss. If both source images exhibit high photometric loss at a particular location, it is highly likely that this area has mismatched points due to dynamic objects, occlusions, or lighting changes. As shown in Fig. \ref{overviewall}, the person riding the bicycle is part of a dynamic region, which exhibits high loss in this area, making it an ideal candidate for masking.

Given the reprojection losses \( \mathbf{L}_{\text{reproj}} \in \mathbb{R}^{B \times C \times H \times W} \), where channel \( c \in \{1, 2\} \) corresponds to the two different reprojection losses. For each channel, we flatten the spatial dimensions into a vector:
\begin{equation}
\mathbf{L}_{c} = \text{Flatten}(\mathbf{L}_{\text{reproj}}[:, c, :, :]) \in \mathbb{R}^{B \times (H \times W)}.
\end{equation}
We then compute the \( p \)-th quantile \( q_c \) for each channel, controlled by a threshold \( \beta \), to identify high-loss regions:
\begin{equation}
q_c = Q_\beta(\mathbf{L}_{c}).
\end{equation}
A higher \( \beta \) excludes only the highest loss areas, while a lower \( \beta \) captures more potential error regions. Using the quantile, a binary mask is generated for each channel:
\begin{equation}
\mathbf{M}_{c} = \mathbb{I}(\mathbf{L}_{\text{reproj}}[:, c, :, :] > q_c),
\end{equation}
and the final dynamic mask is constructed as:
\begin{equation}
\mathbf{M}_{\text{dynamic}} = 1 - (\mathbf{M}_{1} \land \mathbf{M}_{2}).
\end{equation}

\begin{table*}[htbp]
\caption{Depth estimation results on kitti eigen split. Here, ``Test Frames'' indicates the number of frames used for inference, while ``$\bullet$'' denotes whether a semantic segmentation model is utilized.}
\centering
\resizebox{\textwidth}{!}{%
\begin{tabular}{l|c|c|c|cccc|ccc}
\toprule
\multirow{2}{*}{Method} & \multirow{2}{*}{Test Frames} & \multirow{2}{*}{Semantic} & \multirow{2}{*}{W$\times$H} & \multicolumn{4}{c|}{Errors $\downarrow$} & \multicolumn{3}{c}{Accuracy $\uparrow$} \\ \cmidrule(lr){5-8} \cmidrule(lr){9-11}
                        &                             &                           &                              & AbsRel & SqRel & RMSE  & RMSE$_{\log}$ & $\delta < 1.25$ & $\delta < 1.25^2$ & $\delta < 1.25^3$ \\ \midrule
Struct2Depth \cite{stuct2depth}   & 1  & $\bullet$  & 416$\times$128   & 0.141  & 1.026 & 5.291 & 0.215         & 0.816           & 0.945             & 0.979             \\
Bian et al. \cite{bian2019unsupervised} &1& & 416$\times$128          & 0.137  & 1.089 & 5.439 & 0.217 & 0.830& 0.942& 0.975             \\
Gordon et al. \cite{gordon2019depth}  & 1                           & $\bullet$                 & 416$\times$128               & 0.128  & 0.959 & 5.230 & 0.212         & 0.845           & 0.947             & 0.976             \\
Monodepth2 \cite{monodepth2}     & 1                           &                           & 640$\times$192               & 0.115  & 0.903 & 4.863 & 0.193         & 0.877           & 0.959             & 0.981             \\
ManyDepth \cite{Manydepth}     & 2(-1,0)                     &                 & 640$\times$192               & 0.098  & 0.770 & 4.459 & 0.176         & 0.900           & 0.965             & 0.983             \\ 

DynamicDepth \cite{DynamicDepth}    & 2(-1,0)                     & $\bullet$ & 640$\times$192 & 0.096  & 0.720 & 4.458 & 0.175& 0.897 & 0.964  & 0.984 \\
DepthFormer \cite{depthformer} & 2(-1,0) & & 640$\times$192 & 0.090  & 0.661 & 4.149 & 0.175 & 0.905 & 0.967& 0.984 \\ 
DualRefine \cite{DualRefine}     & 2(-1,0)                     &                 & 640$\times$192               & 0.090  & 0.658 & 4.237 & 0.171         & \underline{0.912}           & 0.967             & 0.984             \\ 

MOVEDepth \cite{movedepth}     & 2(-1,0)                     &                 & 640$\times$192               & 0.089  & 0.663 & 4.216 & 0.169         & 0.904           & 0.966             & 0.984             \\

MonoViFI \cite{Monovifi}    & 3(-1,0,1)   &  & 640$\times$192
& 0.091  & \textbf{0.589}  & \underline{4.088} & 0.166 \ & 0.912 \ & \textbf{0.969} \ & \textbf{0.986} \\

\hline

Baseline \cite{xiang2023exploring}  & 2(-1,0) & & 640$\times$192 
& \underline{0.086}  & 0.613 & 4.096 & \underline{0.165} & \underline{0.915} & \textbf{0.969} & \underline{0.985} \\

\textbf{Ours}   & 2(-1,0) & & 640$\times$192 
& \textbf{0.084}  & \underline{0.607} & \textbf{4.086} & \textbf{0.164} & \textbf{0.917} & \textbf{0.969} & \underline{0.985} \\

\bottomrule
\end{tabular}%
}
\label{kitti}
\VspaceS
\end{table*}

\begin{table*}[htbp]
\caption{Depth estimation results on cityscapes. Here, ``Test Frames'' indicates the number of frames used for inference, while ``$\bullet$'' denotes whether a semantic segmentation model is utilized.}
\centering
\resizebox{\textwidth}{!}{%
\begin{tabular}{l|c|c|c|cccc|ccc}
\toprule
\multirow{2}{*}{Method} & \multirow{2}{*}{Test Frames} & \multirow{2}{*}{Semantic} & \multirow{2}{*}{W$\times$H} & \multicolumn{4}{c|}{Errors $\downarrow$} & \multicolumn{3}{c}{Accuracy $\uparrow$} \\ \cmidrule(lr){5-8} \cmidrule(lr){9-11}
                        &                             &                           &                              & AbsRel & SqRel & RMSE  & RMSE$_{\log}$ & $\delta < 1.25$ & $\delta < 1.25^2$ & $\delta < 1.25^3$ \\ \midrule
Struct2Depth \cite{Struct2Depth}   & 1                           & $\bullet$                 & 416$\times$128               & 0.145  & 1.737 & 7.280 & 0.205         & 0.813           & 0.942             & 0.976             \\
Monodepth2 \cite{monodepth2}     & 1                           &                           & 416$\times$128               & 0.129  & 1.569 & 6.876 & 0.187         & 0.849           & 0.957             & 0.983             \\
Gordon et al. \cite{gordon2019depth}  & 1                           & $\bullet$                 & 416$\times$128               & 0.127  & 1.330 & 6.960 & 0.195         & 0.830           & 0.947             & 0.981             \\

ManyDepth \cite{Manydepth}     & 2(-1,0)                     &                 & 416$\times$128              & 0.114  & 1.193 & 6.223 & 0.170         & 0.875           & 0.967             & 0.989             \\

IterDepth \cite{IterDepth}     & 2(-1,0) &   & 416$\times$128               
& 0.104 & 1.032 & 5.724 & 0.157 & 0.896 & \textbf{0.976} & \underline{0.991}  \\

DynamicDepth \cite{DynamicDepth}    & 2(-1,0)   & $\bullet$  & 416$\times$128
& 0.103  & \underline{1.000}  & 5.867 & 0.175 \ & 0.895 \ & 0.967\ & \underline{0.991} \\

MonoViFI \cite{Monovifi}    & 3(-1,0,1)   &  & 416$\times$128
& 0.102  & \textbf{0.836}  & \textbf{5.395} & \underline{0.153} \ & 0.885 \ & 0.971 \ & \textbf{0.992} \\

DS-Depth \cite{DS-Depth}     & 2(-1,0)  &  & 416$\times$128 
& \underline{0.100}  & 1.055 & 5.884 & 0.155 & \underline{0.899}  & \underline{0.974}  & \underline{0.991} \\ 

\hline

Baseline \cite{xiang2023exploring}  & 2(-1,0) & & 416$\times$128  
& 0.102  & \underline{1.000} & 5.792 & 0.156 & 0.892 & 0.973 & \underline{0.991} \\ 

\textbf{Ours}     & 2(-1,0)  &   & 416$\times$128
& \textbf{0.097} & 1.041 & \underline{5.652} & \textbf{0.150} & \textbf{0.909} & \textbf{0.976} & \underline{0.991} \\

\bottomrule
\end{tabular}
}
\label{cityscapes}
\VspaceL
\end{table*}

\subsection{Cost Volume Auto-Masking Strategy}
\label{subsec:CVAM}
Multi-frame depth estimation typically regresses depth maps by constructing a cost volume. However, dynamic objects can adversely affect the accurate construction of this cost volume. Inspired by the Auto-Masking approach in Monodepth2 \cite{monodepth2}, we propose a Cost Volume Auto-Masking strategy for cost volume. This method similarly filters out pixels whose appearance remains unchanged between consecutive frames, meaning that points stationary relative to the camera are masked out.

Given two adjacent frames \(\mathbf{I}_t\) and \(\mathbf{I}_{t-1}\) with pixel values \(\mathbf{I}_t(u, v)\) and \(\mathbf{I}_{t-1}(u, v)\), we define the consistency mask \(\mathbf{M}_{\text{eq}}\) as follows:
\begin{equation}
\mathbf{M}_{\text{eq}}(u, v) =
\begin{cases}
1 & \text{if } \mathbf{I}_t(u, v) = \mathbf{I}_{t-1}(u, v), \\
0 & \text{otherwise}.
\end{cases}
\end{equation}
The cost volume auto-masking mask \(\mathbf{M}_{\text{cost}}\) is derived as:
\begin{equation}
\mathbf{M}_{\text{cost}}(u, v) = 1 - \mathbf{M}_{\text{eq}}(u, v).
\end{equation}
This mask is zero when all channels are consistent and one otherwise. 
To apply this mask, it is downsampled to match the resolution of the feature maps. Let \( s = 2^{\text{scale}} \) be the downsampling factor. The downsampled mask \(\mathbf{M}_{\text{down}}\) is obtained by:
\begin{equation}
\mathbf{M}_{\text{down}}(u', v') = \mathbf{M}_{\text{cost}}\left(\left\lfloor \frac{u}{s} \right\rfloor, \left\lfloor \frac{v}{s} \right\rfloor \right).
\end{equation}

This downsampled mask is then applied element-wise to the feature maps of the reference frame \(\mathbf{F}_{\text{ref}}\) and the source frame \(\mathbf{F}_{\text{source}}\) as follows:
\begin{equation}
\mathbf{F}_{\text{ref}}'(u', v') = \mathbf{F}_{\text{ref}}(u', v') \odot \mathbf{M}_{\text{down}}(u', v'),
\end{equation}
\begin{equation}
\mathbf{F}_{\text{source}}'(u', v') = \mathbf{F}_{\text{source}}(u', v') \odot \mathbf{M}_{\text{down}}(u', v').
\end{equation}
Here, \(\odot\) denotes element-wise multiplication. 
The benefit of this approach is that it allows for the exclusion of objects moving at the same speed as the camera before constructing the cost volume. It even enables the omission of entire frames from monocular video when the camera is stationary. This strategy diverges from Monodepth2 \cite{monodepth2} by specifically addressing photometric inconsistencies caused by occlusions and motion when constructing the cost volume. Furthermore, this approach lays the groundwork for the subsequent Spectral Entropy Uncertainty Module and guides its implementation.

\subsection{Spectral Entropy Uncertainty Module}
\label{subsec:SEU}
To address dynamic object issues and enhance depth regression using cost volume information, we propose the Spectral Entropy Uncertainty (SEU) module. This module integrates spectral entropy with Fourier transform for uncertainty estimation and depth fusion. The Fourier transform converts spatial information into the frequency domain, facilitating the identification of noise introduced by dynamic objects. And spectral entropy analysis quantifies the complexity of these frequency components, allowing for a more precise characterization and management of uncertainty regions.
Firstly, we compute the Fourier transform of the cost probability tensor \( C \):
\begin{equation}
\hat{C} = FFT(C).
\end{equation}

We then derive the magnitude spectrum \( S \) and normalize it to obtain a probability distribution \( P \):
\begin{equation}
M = |\hat{C}|, \quad P = \frac{S}{\sum S}.
\end{equation}

Next, we calculate the spectral entropy \( H \) from \( P \):
\begin{equation}
H = -\sum P \cdot \log(P + \epsilon).
\end{equation}
Following this, we use a neural network to predict the uncertainty \( U \) from \( H \).

Finally, we fuse the depth estimates \( D_{multi} \) and \( D_{mono} \) using the predicted uncertainty \( U \) to obtain the fused depth \( D_f \) following MOVEDepth \cite{movedepth}:
\begin{equation}
D_{Fuse} = (1 - U) \cdot D_{multi} + U \cdot D_{mono}.
\end{equation}

The proposed strategy integrates single-frame and multi-frame depth estimates, dynamically adjusting the fusion process based on the Spectral Entropy Uncertainty Module. Additionally, the fused depth $D_{Fuse}$ is used solely for loss computation, thereby not introducing additional inference overhead.

\subsection{Training Objective}
Our D\textsuperscript{3}epth is trained in a self-supervised manner, and the loss consists of three parts:
\begin{equation}
\mathcal{L}_{\text{total}} = \mathcal{L}_{\text{Mono}}(D_{\text{Mono}}) + \mathcal{L}_{\text{Multi}}(D_{\text{Multi}}) + \mathcal{L}_{\text{Fuse}}(D_{\text{Fuse}}),
\end{equation}
where \( \mathcal{L}_\text{Mono} \), \( \mathcal{L}_\text{Multi} \), and \( \mathcal{L}_\text{Fuse} \) represent the loss associated with the depth estimates \( D_\text{Mono} \), \( D_\text{Multi} \), and \( D_\text{Fuse} \), respectively. And \(\mathcal{L}(\cdot)\) is a weighted combination of reprojection loss \( \mathcal{L}_{ph} \) (Eq. \ref{eq:photometric}) and depth smooth loss \( \mathcal{L}_{s} \) \cite{godard2017unsupervised}:
\begin{equation}
\mathcal{L}(D) = \mathcal{L}_{ph}(D) \cdot \mathbf{M}_{\text{dynamic}} + \gamma \mathcal{L}_{s}(D),
\end{equation}
where \(\gamma = 0.001\) denotes the loss weight.

\section{EXPERIMENTS}

\subsection{Dataset}
We evaluate our approach on two widely-used benchmarks for depth estimation: KITTI \cite{kitti} and Cityscapes \cite{cityscapes}. KITTI serves as a standard dataset for outdoor driving scenarios, providing diverse scenes. Following the Eigen split \cite{eigensplit}, we use 39,810 images for training, 4,424 for validation, and 697 for testing. The Cityscapes dataset features a higher proportion of dynamic objects, making it particularly well-suited for evaluating our method’s performance in dynamic environments. We train on 69,731 monocular triplets and evaluate on 1,525 test images. Since our method is specifically designed to address the challenges posed by dynamic objects, and considering that the Cityscapes dataset features a high prevalence of moving objects, the majority of our experiments are conducted on Cityscapes. 
Both datasets are evaluated with a maximum depth of 80 meters.

\subsection{Implementation Details}

In our experiments, we implement the models using PyTorch and train them on NVIDIA RTX 4090 GPU. Our D\textsuperscript{3}epth is based on a two-stage teacher-student distillation method following \cite{xiang2023exploring}. The dynamic mask threshold \(\beta\) is set to 0.8. For the KITTI dataset, this threshold is applied starting from epoch 0, while for the Cityscapes dataset, it is applied from epoch 1. We optimize the models using the Adam optimizer \cite{kingma2014adam}, with an initial learning rate of $2 \times 10^{-4}$ for the teacher model and $1 \times 10^{-4}$ for the student model. The learning rate is reduced by a factor of 10 after a specific number of epochs: 15 epochs for KITTI and 1 epoch for Cityscapes. The total number of training epochs is 20 for KITTI and 5 for Cityscapes, with a batch size of 12.

\begin{table}[tpb]
\caption{Ablation study on the different components of our D\textsuperscript{3}epth on the Cityscapes dataset. Here, DM represents the Dynamic Mask, CAM refers to the Cost Volume Auto-Masking strategy, and SEU denotes the Spectral Entropy Uncertainty Module.}
\centering
\renewcommand{\arraystretch}{1.1} 
\setlength{\tabcolsep}{3pt} 
\begin{tabular}{|l||>{\centering\arraybackslash}p{1.1cm}|>{\centering\arraybackslash}p{1.1cm}|>{\centering\arraybackslash}p{1.1cm}|>{\centering\arraybackslash}p{1.1cm}|} 
\hline
\textbf{Method} & 
\cellcolor[HTML]{F4CCCC} \textbf{Abs Rel} & 
\cellcolor[HTML]{F4CCCC} \textbf{RMSE$_{\log}$} & 
\cellcolor[HTML]{C9DAF8} \boldmath{$\delta_1$} & 
\cellcolor[HTML]{C9DAF8} \boldmath{$\delta_2$} \\ \hline

\multicolumn{5}{|c|}{\cellcolor[HTML]{FFF2CC} \textbf{Single-Frame Inference}} \\ \hline
Baseline & 0.105 & 0.158 & 0.891 & 0.972 \\ 
Baseline + DM & 0.104 & 0.156 & 0.899 & 0.974 \\ 
Baseline + CAM & \textbf{0.101} & 0.156 & 0.890 & 0.973 \\ 
Baseline + SEU & 0.102 & \textbf{0.155} & 0.896 & 0.973 \\ 
Baseline + DM + CAM & 0.103 & \textbf{0.155} & 0.898 & 0.974 \\ 
\textbf{D\textsuperscript{3}epth (Single-frame)} & 0.102 & \textbf{0.155} & \textbf{0.904} & \textbf{0.975} \\ \hline

\multicolumn{5}{|c|}{\cellcolor[HTML]{D9EAD3} \textbf{Multi-Frame Inference}} \\ \hline
Baseline & 0.102 & 0.156 & 0.892 & 0.973 \\
Baseline + DM & 0.100 & 0.152 & 0.904 & 0.975 \\ 
Baseline + CAM & 0.100 & 0.155 & 0.890 & 0.973 \\ 
Baseline + SEU & 0.101 & 0.154 & 0.896 & 0.973 \\ 
Baseline + DM + CAM & 0.099 & 0.151 & 0.904 & 0.975 \\

\textbf{D\textsuperscript{3}epth (Multi-frame)} & \textbf{0.097} & \textbf{0.150} & \textbf{0.909} & \textbf{0.976} \\ \hline

\end{tabular}
\label{ablation}
\VspaceL
\end{table}

\subsection{Comparison with State-of-the-Art Methods}
We evaluate our method on the KITTI and Cityscapes benchmarks, with the results presented in Table \ref{kitti} and Table \ref{cityscapes}. As reported in previous studies \cite{DynamicDepth, MAL}, dynamic objects such as vehicles, pedestrians, and cyclists make up only 0.34\% of the pixels in the KITTI dataset, with the majority of vehicles being stationary. Therefore, our method achieves only a marginal improvement on the KITTI dataset, while still reaching state-of-the-art performance. Specifically, $\delta < 1.25$ improves by 0.002. However, on the Cityscapes dataset, which features more prevalent dynamic scenes, our method exhibits a significant improvement. We observe an increase of 0.017 in $\delta < 1.25$, and the absolute relative error (Abs Rel) decreases to 0.087, substantially outperforming existing methods designed to handle dynamic objects. 

\begin{figure}[tbp]
\centering
\includegraphics[width=1.00\linewidth]{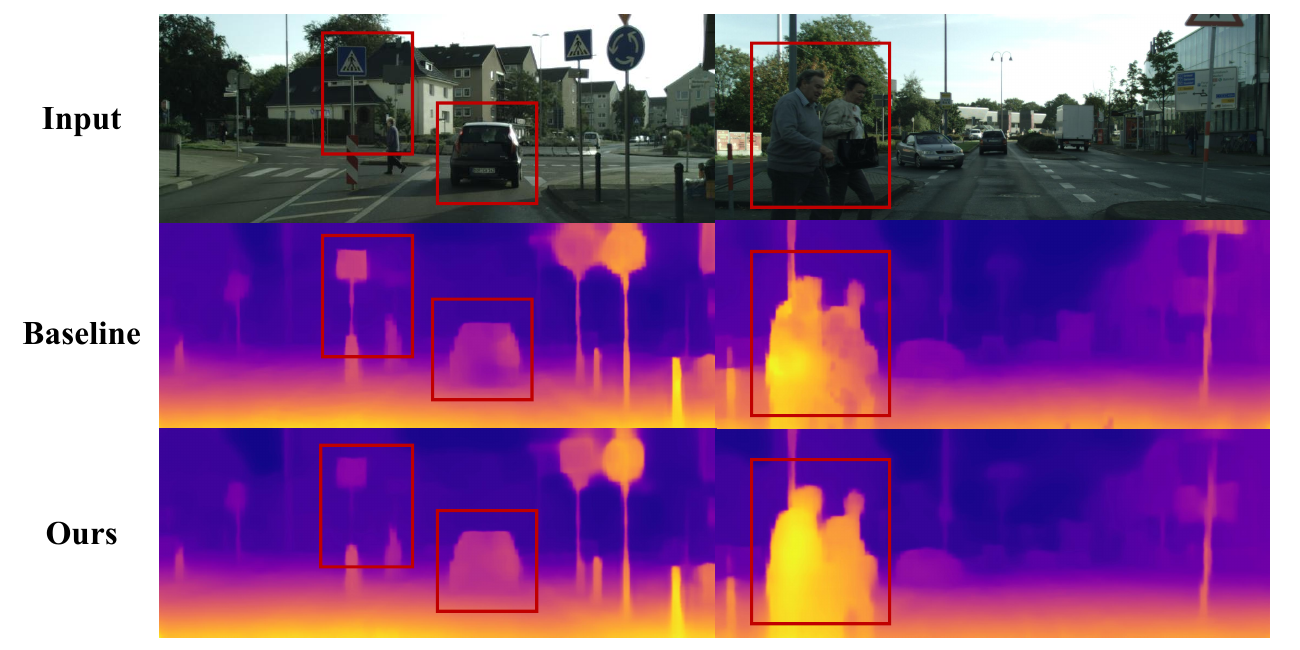}
\caption{Qualitative results of D\textsuperscript{3}epth and the baseline on Cityscapes dataset. The first row displays the input images, the second row shows the baseline results, and the third row presents the outputs of our D\textsuperscript{3}epth. }
\label{visual}
\VspaceL
\end{figure}

We also conduct visualization experiments on the Cityscapes dataset. The qualitative results are presented in Fig. \ref{visual}. Specifically designed to tackle challenges related to dynamic objects, our method demonstrates enhanced accuracy in areas containing moving vehicles and pedestrians, as indicated by the red boxes. Additionally, it offers a more precise delineation of fine details, such as the road signs highlighted in red.

\subsection{Ablation Study}

In this section, we discuss the impact of each component within our D\textsuperscript{3}epth. Our ablation experiments are primarily conducted on the Cityscapes dataset.

\noindent \textbf{Impact of Different Components of D\textsuperscript{3}epth.} First, we perform ablation studies on the single-frame model, as shown in the top part of Table \ref{ablation}. The results indicate that each component (DM, CAM, SEU) contributes to accuracy improvements. Moreover, combining these components yields additional performance gains. When all elements are integrated, our single-frame model achieves the best performance, surpassing even the current state-of-the-art multi-frame methods, thus validating the effectiveness of these modules in addressing dynamic object issues.

Next, we conduct ablation experiments under multi-frame inference, as shown in the bottom part of Table \ref{ablation}. The results demonstrate that each component contributes to performance improvements, and combining these strategies leads to even greater gains. This validates the effectiveness of our approach: the CAM strategy provides early-stage guidance for identifying dynamic objects, which in turn informs the SEU module. Simultaneously, the DM method enforces global constraints within a self-supervised framework. The synergy of these components results in state-of-the-art performance.

\noindent \textbf{Effectiveness of FFT strategy in SEU.}
The Fourier transform converts spatial information into the frequency domain, making it easier to identify noise introduced by dynamic objects. We validate its effectiveness through ablation experiments on the Cityscapes dataset, as shown in Table \ref{table:fft_ablation}. The results demonstrate that the use of the FFT technique leads to notable performance improvements. 

\begin{table}[tpb]
\caption{Ablation study on the use of FFT strategy in the
Spectral Entropy Uncertainty Module.}
\centering
\renewcommand{\arraystretch}{1.1} 
\setlength{\tabcolsep}{3pt} 
\begin{tabular}{|l||>{\centering\arraybackslash}p{1.1cm}|>{\centering\arraybackslash}p{1.1cm}|>{\centering\arraybackslash}p{1.1cm}|>{\centering\arraybackslash}p{1.1cm}|} 
\hline
\textbf{Method} & 
\cellcolor[HTML]{F4CCCC} \textbf{Abs Rel} & 
\cellcolor[HTML]{F4CCCC} \textbf{RMSE$_{\log}$} & 
\cellcolor[HTML]{C9DAF8} \boldmath{$\delta_1$} & 
\cellcolor[HTML]{C9DAF8} \boldmath{$\delta_2$} \\ \hline
D\textsuperscript{3}epth w/o FFT& 0.099 & 0.153 & 0.905 & 0.975 \\
D\textsuperscript{3}epth & \textbf{0.097} & \textbf{0.150} & \textbf{0.909} & \textbf{0.976} \\ 
\hline
\end{tabular}
\label{table:fft_ablation}
\VspaceS
\end{table}

\section{CONCLUSION}
In this paper, we propose D\textsuperscript{3}epth, a novel self-supervised depth estimation framework specifically designed to tackle challenges posed by dynamic objects. We introduce the Dynamic Mask to mitigate the impact of dynamic objects on the loss function, improving robustness in handling dynamic entities. Additionally, we enhance multi-frame depth estimation using the Cost Volume Auto-Masking strategy, which identifies and masks potential dynamic object regions. Furthermore, we incorporate the Spectral Entropy Uncertainty module to guide depth fusion and address challenges associated with dynamic objects in the cost volume. Our method achieves state-of-the-art results on the KITTI and Cityscapes datasets. Future work will focus on refining the distinction between high-loss areas caused by dynamic objects and those inherently exhibiting high loss, to more accurately localize dynamic objects within the reprojection loss. 

\bibliographystyle{IEEEtran}
\bibliography{ref.bib}
\end{document}